\def\year{2020}\relax
\documentclass[letterpaper]{article} 
\usepackage{aaai20}  
\usepackage{times}  
\usepackage{helvet} 
\usepackage{courier}  
\usepackage[hyphens]{url}  
\usepackage{graphicx} 
\usepackage{pdfpages}
\usepackage{graphicx}  
\usepackage[utf8]{inputenc}

\usepackage{subcaption}
\usepackage{amsmath}
\usepackage{amssymb}
\usepackage{url}
\usepackage{array}
\newcolumntype{L}[1]{>{\raggedright\let\newline\\\arraybackslash\hspace{0pt}}m{#1}}
\newcolumntype{C}[1]{>{\centering\let\newline\\\arraybackslash\hspace{0pt}}m{#1}}
\usepackage{multirow}
\usepackage[usestackEOL]{stackengine}

\def\bx{\bs{x}}
\def\bz{\bs{z}}

\def\bu{\bs{u}}

\def\bss{\bs{s}}

\def\bx{\bs{x}}
\def\bz{\bs{z}}

\def\bw{\bs{w}}
\def\bC{\bs{C}}
\def\bB{\bs{B}}

\def\bX{\bs{X}}
\def\bbR{\mathbb{R}}

\def\pola{A} 
\def\pols{S} 
\newcommand{\Figure}[1]{Fig.~\ref{#1}}
\newcommand{\Table}[1]{Table~\ref{#1}}
\newcommand{\Section}[1]{\S\ref{#1}}
\newcommand{\bs}{\boldsymbol}

\newcommand{\pith}{\pi_{\bw}}
\newcommand{\piw}{\pi_{\bw}}

\newcommand{\V}[1]{\mathcal{V}^{#1}}
\newcommand{\Vt}{\mathcal{V}_t}

\newcommand{\Sone}{\mathcal{S}^{1 \colon T}}
\newcommand{\sigm}[1]{\sigma\left[ #1 \right]}

\title{Deep Reinforcement Learning for Active Human Pose Estimation}
\author{\Large \textbf{Erik G\"artner\textsuperscript{\rm 1}\thanks{Denotes equal contribution, order determined by coin flip.}, Aleksis Pirinen\textsuperscript{\rm 1}\textsuperscript{\rm}\footnotemark[1], Cristian Sminchisescu\textsuperscript{\rm 1,}\textsuperscript{\rm 2}}\\ 
\textsuperscript{\rm 1}Department of Mathematics, Faculty of Engineering, Lund University\\
\textsuperscript{\rm 2}Google Research\\
\{erik.gartner, aleksis.pirinen, cristian.sminchisescu\}@math.lth.se
}
 
\begin{document}

\maketitle

\begin{abstract}
Most 3d human pose estimation methods assume that input -- be it images of a scene collected from one or several viewpoints, or from a video -- is given. Consequently, they focus on estimates leveraging prior knowledge and measurement by fusing information spatially and/or temporally, whenever available. In this paper we address the problem of an active observer with freedom to move and explore the scene spatially -- in `time-freeze' mode -- and/or temporally, by selecting informative viewpoints that improve its estimation accuracy. Towards this end, we introduce \emph{Pose-DRL}, a fully trainable deep reinforcement learning-based active pose estimation architecture which learns to select appropriate views, in space and time, to feed an underlying monocular pose estimator. We evaluate our model using single- and multi-target estimators with strong result in both settings. Our system further learns automatic stopping conditions in time and transition functions to the next temporal processing step in videos. In extensive experiments with the Panoptic multi-view setup, and for complex scenes containing multiple people, we show that our model learns to select viewpoints that yield significantly more accurate pose estimates compared to strong multi-view baselines. Code is available: \url{https://github.com/aleksispi/pose-drl}.
\end{abstract}

\section{Introduction}
Existing human pose estimation models, be them designed for 2d or 3d reconstruction, assume that viewpoint selection is outside the control of the estimation agent. This problem is usually solved by a human, either once and for all, or by moving around and tracking the elements of interest in the scene. Consequently, the work is split between \emph{sufficiency} (e.g. instrumenting the space with as many cameras as possible in motion capture setups), \emph{minimalism} (work with as little as possible, ideally a single view, as given), or \emph{pragmatism} (use whatever is available, e.g. a stereo system and lidar in a self-driving car). While each of these scenarios and their underlying methodologies make practical or conceptual sense in their context of applicability, none covers the case of an active observer moving in the scene in order to reduce uncertainty, with emphasis on trading accuracy and computational complexity. There are good reasons for this, as experimenting with an active system faces the difficulty of linking perception and action in the real world, or may have to resort on simulation, which can however lack visual appearance and motion realism, especially for complex articulated and deformable structures such as people. 

In this work we consider 3d human pose estimation from the perspective of an active observer,
and operate with an idealization that allows us to distill the active vision concepts, develop new methodology, and test it on real image data. We work with a Panoptic massive camera grid \cite{Joo_2015_ICCV}, where we can both observe the scene in time-freeze, from a dense variety of viewpoints, and process the scene temporally, thus being able to emulate a moving observer. An active setup for 3d human pose estimation addresses the incomplete body pose observability in any monocular image due to depth ambiguities or occlusions (self-induced or produced by other people or objects). It also enables adaptation with respect to any potential limitations of the associated pose estimation system, by sequentially selecting views that together yield accurate pose predictions.

In this context we introduce \emph{Pose-DRL}, a deep reinforcement learning (RL) based active pose estimation architecture operating in a dense camera rig, which learns to select appropriate viewpoints to feed an underlying monocular pose predictor. Moreover, our model learns when to stop viewpoint exploration in time-freeze, or continue to the next temporal step when processing video. In evaluations using Panoptic we show that our system learns to select sets of views yielding more accurate pose estimates compared to strong multi-view baselines. The results not only show the advantage of intelligent viewpoint selection, but also that often `less is more', as fusing too many possibly incorrect viewpoint estimates leads to inferior results.

As our model consists of a deep RL-based active vision module on top of a task module, it can be easily adjusted for other visual routines in the context of a multi-camera setup by simply replacing the task module and retraining the active vision component, or refining them jointly in case of access and compatibility. We show encouraging results using different pose estimators and task settings.

\section{Related Work}
Extracting 2d and 3d human representations from \emph{given} images or video is a vast research area, recently fueled by progress in keypoint detection \cite{wei2016convolutional,Papandreou_2017_CVPR}, semantic body parts segmentation \cite{popa2017deep}, 3d human body models \cite{SMPL2015}, and 3d motion capture data \cite{h36m_pami,vonMarcard2018}. Deep learning plays a key role in most human pose and shape estimation pipelines \cite{bogo2016,rhodin2016general,popa2017deep,pavlakos2017cvpr,rogez2017lcr,zanfir2018monocular,mehta2017vnect,kanazawa2018end}, sometimes in connection with non-linear refinement \cite{bogo2016,zanfir2018monocular}. Systems integrating detailed face, body and hand models have also been proposed \cite{totalcapture_cvpr18}. Even so, the monocular 3d case is challenging due to depth ambiguities which motivated the use of additional ordering constraints during training \cite{pavlakos2018ordinal}.

In addition to recent literature for static pipelines, the community has recently seen an increased interest for active vision tasks, including RL-based visual navigation \cite{ammirato2017dataset,das2018embodied,xia2018gibson,zhu2017target}. In \cite{ammirato2017dataset} a real-world dataset of sampled indoor locations along multiple viewing directions is introduced. An RL-agent is trained to navigate to views in which a given instance detector is accurate, similar in spirit to what we do, but in a different context and task.

A joint gripping and viewing policy is introduced in \cite{cheng2018reinforcement}, also related to us in seeking policies that choose occlusion-free views. The authors of \cite{cheng2018geometry} introduce an active view selection system and jointly learn a geometry-aware model for constructing a 3d feature tensor, which is fused together from views predicted by a policy network. In contrast to us, their policy predicts one of 8 adjacent discrete camera locations, they do not consider moving objects, their model does not automatically stop view selection, and they do not use real data. In \cite{jayaraman2018learning,xiong2018snap}, active view selection is considered for panoramic completion and panorama projection, respectively. Differently from us, their view selection policies operate on discretized spheres and do not learn automatic stopping conditions. An approach for active multi-view object recognition is proposed in \cite{johns2016pairwise}, where pairs of images in a view trajectory are sequentially fed to a CNN for recognition and for next best view prediction. Optimization is done over discretized movements and pre-set trajectory lengths, in contrast to us.

Most related to us is \cite{NIPS2019_8646}, who also consider active view selection in the context of human pose estimation. However, they work with 2d joint detectors and learn to actively triangulate those into 3d pose reconstructions. Thus we face different challenges -- while \cite{NIPS2019_8646} only require each joint to be visible in two views for triangulation, our model has to consider which views yield accurate fused estimates. Furthermore, their model does not learn a stopping action that trades accuracy for speed, and they do not study both the single-target and multi-target cases, as we do in this paper. 

Aside from active vision applications in real or simulated environments, reinforcement learning has also been successfully applied to other vision tasks, e.g. object detection \cite{CaicedoLazebnik-iccv-2015,pirinen2018deep}, object tracking \cite{zhang2017deep,yun2018action} and visual question answering \cite{das2017learning}.
\section{Active Human Pose Estimation}\label{sec:act_pose_pred}

\begin{figure*}[t]
    \begin{center}
     \includegraphics[scale=0.5]{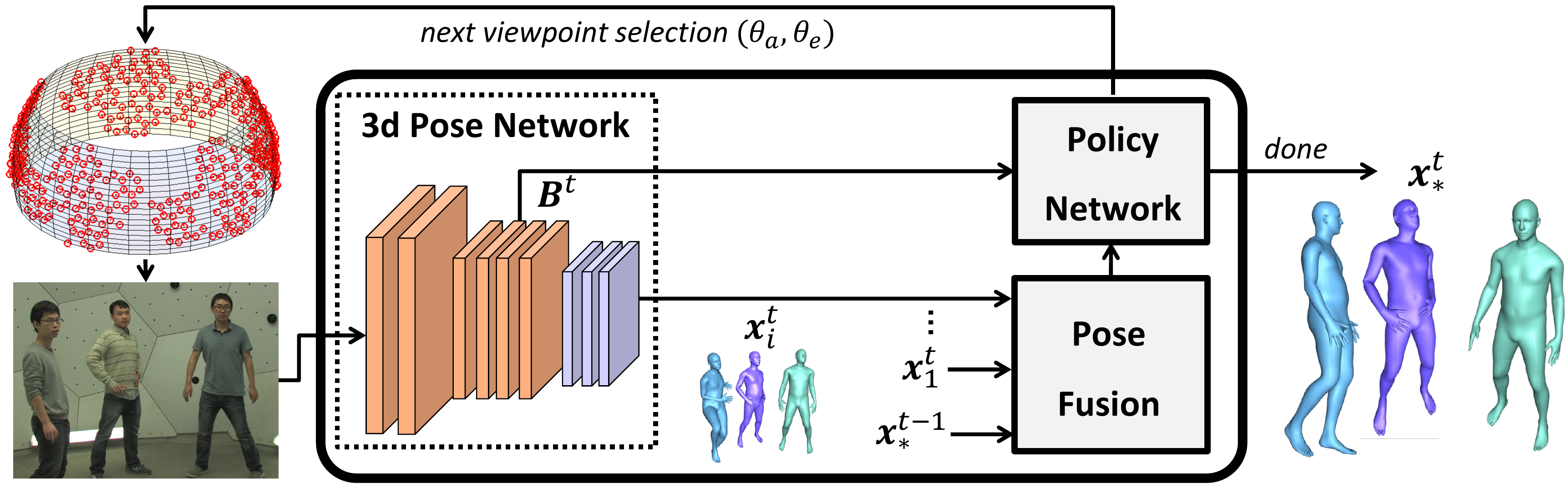}
    \caption{Overview of our Pose-DRL agent for active human pose estimation. The agent initially observes the scene from a randomly given camera on the rig. In each visited viewpoint, the associated image is processed by a 3d pose estimation network, producing the base state $\bB^t$ of the agent and pose estimate(s) $\bx^t_{i}$. The pose estimate is fused together with estimates from previous viewpoints $\bx^t_{1}, \dots, \bx^t_{i-1}$ and the previous temporal step $\bx_\star^{t-1}$. Both the current and fused estimate are fed as additional features to the agent. At each step the policy network outputs the next viewpoint, until it decides it is done and continues to next active-view at time $t+1$. The viewpoint selection action predicts spherical angles relative to the agent's current location on the camera rig, and the closest camera associated with the predicted angles is visited next. When the agent is done it outputs $\bx^t_\star$, the per-joint fusion of the individual pose estimates seen during the current active-view and the fused pose estimate from the previous active-view, cf. \eqref{eq:pose_fusion}. Pose-DRL can be used either to reconstruct a target person, or to reconstruct all people in the scene. The underlying pose estimator is exchangeable -- we show strong results using two different ones in \Section{sec:main-results}.}\label{fig:overview-one-mc}
    \end{center}
\end{figure*}

In this section we describe our active human pose estimation framework, arguing it is a good proxy for a set of problems where an agent has to actively explore to understand the scene and integrate task relevant information. For example, a single view may only contain parts of the human body (or be absent of the person altogether) and the agent needs to find a better view to capture the person's pose. Pose estimators are often trained on a limited set of viewing angles and yield lower performance for others. Our setup forces the agent to also take any estimation limitations into account when selecting multiple views. In particular, we show in \Section{sec:main-results} that learning to find good views and fusing them is more important than relying on a large number of random ones, or the full available set, as standard -- see also \Figure{fig:err-all-cams}.

Concepts in the following sections will, for simplicity, be described assuming the model is estimating the pose of a single \emph{target person} (though scenes may contain multiple people occluding the target). The setting in which \emph{all} people are reconstructed simultaneously is described in \Section{sec:multiple-people}.

\subsection{Active Pose Estimation Setup}\label{sec:active-setup}
We idealize our active pose estimation setup using CMU's Panoptic installation \cite{Joo_2015_ICCV} as it captures real video data of scenes with multiple people and cameras densely covering the viewing sphere. This allows us to simulate an active agent observing the scene from multiple views, without the complexity of actually moving a camera. It also enables controllable and reproducible experiments. The videos are captured in a large spherical dome fitted with synchronized HD cameras.\footnote{There are about 30 cameras per scene. The HD cameras provide better image quality than VGA and are sufficiently dense, yet spread apart far enough to make each viewpoint unique.} Inside the dome several human actors perform a range of movements, with 2d and 3d joint annotations available. The dataset is divided into a number of \emph{scenes}, video recordings from all synchronized cameras capturing different actors and types of movements, ranging from simple pose demonstrations to intricate social games. \\ \\
\noindent{\bf Terminology.} We call a \emph{time-freeze} $\{v^t_1, \dots, v^t_N\}$ the collection of views from all $N$ time-synchronized cameras at time $t$, with $v^t_i$ the image (referred to as \emph{view} or \emph{viewpoint}) taken by camera $i$ at time $t$. A subset of a time-freeze is an \emph{active-view} $\V{t} = \{v^t_{1}, \ldots, v^t_{k}\}$ containing $k$ selected views from the time-freeze. A temporally contiguous sequence of active-views is referred to as an \emph{active-sequence}, $\Sone = \{\V{1}, \V{2}, \dots, \V{T}\}$. We will often omit the time superfix $t$ unless the context is unclear; most concepts will be explained at the level of time-freezes. The image corresponding to a view $v_i$ can be fed to a pose predictor to produce a pose estimate $\bx_i \in \bbR^{45}$ ($15\times$ 3d joints).\\ \\
\noindent{\bf Task definition.} We define the task of \emph{active pose estimation} at each time step as selecting views from a time-freeze to generate an active-view. The objective is to produce an accurate fused estimate $\bx_\star$ from pose predictions $\bx_{1}, \ldots, \bx_{k}$ associated with the active-view ($k$ may vary between active-views). The deep pose estimation network is computationally demanding and therefore working with non-maximal sets of views decreases processing time. Moreover, it improves estimates by ignoring obstructed views, or those a given pose predictor cannot accurately handle. The goal of the full active pose estimation task is to produce accurate fused pose estimates over the full sequence, i.e., to produce an active-sequence with accurate corresponding fused pose estimates.
\begin{figure*}
    \begin{center}
     \includegraphics[width=0.91\textwidth]{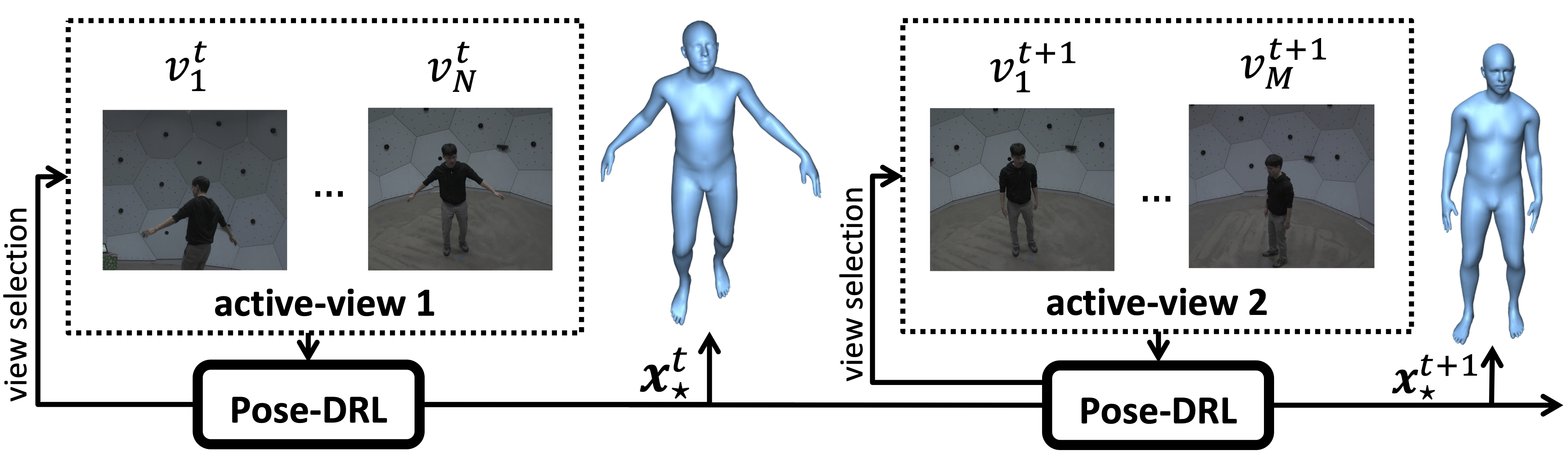}
    \caption{Illustration of how Pose-DRL operates on an active-sequence, here shown for a single-person scenario. Fused pose estimates are fed to subsequent active-views within the active-sequence, both as additional state representation for action selection, and for fusing poses temporally.}\label{fig:overview-sub-scene}
    \end{center}
\end{figure*}
\subsection{Detection and Matching of Multiple People}\label{sec:matching}
To solve active human pose estimation the model must address the problems of detecting, tracking, and distinguishing people in a scene. It must also be robust to variations in appearance since people are observed over time and from different views. We use Faster R-CNN \cite{ren2015faster} for detecting people. At the start of an active-sequence the agent is given appearance models, consisting of instance-sensitive features for each person. For each visited view, the agent computes instance features for all detected persons, comparing them with the given appearance models to identify the different people.  \\
\noindent\textbf{Obtaining appearance models.} A generic instance classifier, implemented as a VGG-19 based  \cite{simonyan15} siamese network, is trained for 40k iterations on the training set with a contrastive loss to distinguish between different persons. Each mini-batch (of size 16) consists of randomly sampled pairs of ground-truth crops of people in the training set. We ensure that the training is balanced by sampling pairs of person crops such that the probability of the two crops containing the same person is the same as that of containing two different persons. The people crops are sampled uniformly across scenes, spatially and temporally, yielding a robust instance classifier. 

Once the instance classifier has been trained, we fine-tune it for 2k iterations for each scene and then use it to construct appearance models at the beginning of an active-sequence. For each person, we sample $L$ instance features from time-freezes from the same scene, but outside of the time span of the current active-sequence to limit overfitting. Denote by $\bu_i^l$, the $i$:th instance feature for the $l$:th person, with $i = 1, \dots, L$. Then we set as appearance model:
\begin{equation}\label{eq:appearance_model}
\bs{m}^l = \text{median}(\bu_1^l,\dots, \bu_L^l)    
\end{equation}
We set $L=10$ to obtain a diverse set of instance features for each person, yielding a robust appearance model. \\ \\
\noindent\textbf{Stable matching of detections.} In each visited viewpoint during an active-sequence, the agent computes instance features for all detected persons, comparing them with the given appearance models to identify the different people. To ensure a stable matching, we use the Hungarian algorithm. Specifically, the cost $c^{j,l}$ of matching the $j$:th detection with instance feature $\bu^j$ in the current viewpoint to the appearance model $\bs{m}^l$ of the $l$:th person is $c^{j,l}=\|\bu^j - \bs{m}^l\|_2^2$.
Since the target person may not be visible in all viewpoints throughout the active-sequence, we specify a \emph{cost threshold}, $\mathcal{C} = 0.5$, such that if the assignment cost $c^{j,l}$ of the target is above it (i.e. $c^{j,l} > \mathcal{C}$), we consider the person to not be visible in the view. In that case the associated pose is not fused into the final estimate.

\section{Deep Reinforcement Learning Model}
We now introduce our Pose-DRL agent for solving the active human pose estimation task and first explain the agent's state representation and actions, then define the reward signal for training an agent which selects views that yield an accurate fused pose estimate while keeping down processing time. 

\subsection{Overview of the Pose-DRL Agent}\label{sec:states_actions}
The agent is initiated at a randomly selected view $v_{1}^1$ in the first active-view $\V{1}$ of an active-sequence $\Sone$. Within the current active-view $\V{t}$, the agent issues \emph{viewpoint selection} actions to progressively select a sequence of views $v_{2}^t, \dots, v_{k}^t$, the number of which may vary between active-views. At each view $v_{i}^t$ the underlying pose estimator predicts the pose $\bx_{i}^t$. As seen in \Figure{fig:overview-one-mc} the cameras are approximately located on a partial sphere, so a viewpoint can be specified by the azimuth and elevation angles (referred to as \emph{spherical angles}). Thus for viewpoint selection the Pose-DRL agent predicts spherical angles relative to its current location and selects the camera closest to those angles.

Once the agent is done exploring viewpoints associated to a particular time-freeze it issues the \emph{continue} action and switches to the next active-view $\V{t+1}$, at which time the collection of individual pose estimates $\bx_{i}^t$ from the different viewpoints are fused together with the estimate $\bx_\star^{t-1}$ from the previous active-view $\V{t-1}$:
\begin{equation}\label{eq:pose_fusion}
\bx_\star^{t} = f(\bx_\star^{t-1}, \bx_{1}^t, \bx_{2}^t, \ldots, \bx_{k}^t)
\end{equation}
Including the previous time step estimate $\bx_\star^{t-1}$ in the pose fusion as in \eqref{eq:pose_fusion} often improves results (see \S\ref{sec:ablations}). After returning the fused pose estimate $\bx_\star^{t}$ for the current active-view, the agent continues to the next active-view $\V{t+1}$. The initial view $v^{t+1}_{1}$ for $\V{t+1}$ is set to the final view $v^t_{k}$ of $\Vt$, i.e., $v^{t+1}_{1} = v^t_{k}$. The process repeats until the end of the active-sequence. Fig. \ref{fig:overview-one-mc} and \ref{fig:overview-sub-scene} show model overviews for active-views and active-sequences, respectively.
\begin{table*}[!htbp]
\begin{center}
\scalebox{0.77}{
\begin{tabular}{|c|c|c|c|c|c|c|c|c|c|c|c|c|c|c|}
\cline{1-7}
\cline{9-15}
\textbf{\Centerstack{Model}} & \textbf{\Centerstack{\# Views}} & \textbf{\Centerstack{Maf}} & \textbf{\Centerstack{Ult}} & \textbf{\Centerstack{Pose}} & \textbf{\Centerstack{Maf + Ult}} & \textbf{\Centerstack{All}} & & \textbf{\Centerstack{Model}} & \textbf{\Centerstack{\# Views}} & \textbf{\Centerstack{Maf}} & \textbf{\Centerstack{Ult}} & \textbf{\Centerstack{Pose}} & \textbf{\Centerstack{Maf + Ult}} & \textbf{\Centerstack{All}} \\
\cline{1-7}
\cline{9-15}
\multirow{4}{2.1cm}{\centering\textbf{Pose-DRL-S}} 
             & \multirow{2}{1.0cm}{\centering\textbf{auto}} & $130.3$ & $135.4$ & $135.3$ & $134.2$ & $135.0$ & & \multirow{4}{2.1cm}{\centering\textbf{Pose-DRL-M}} & \multirow{2}{1.0cm}{\centering\textbf{auto}} & $114.8$ & $116.4$ & $104.6$ & $115.9$ & $110.7$ \\
              & & $(4.6)$ & $(3.4)$ & $(3.7)$ & $(3.8)$ & $(3.7)$  & & &  & $(7.5)$ & $(6.6)$ & $(2.1)$ & $(6.8)$ & $(4.5)$ \\
\cline{2-7}
\cline{10-15}
             & \multirow{2}{1.0cm}{\centering\textbf{fixed}} & $144.7$ & $157.5$ & $135.1$ & $155.5$ & $140.4$ & & & \multirow{2}{1.0cm}{\centering\textbf{fixed}} & 114.8 & 118.0 & 106.7 & 117.6 & 112.8 \\
             &  & (5.0) & (4.0) & (4.0) & (4.0) &  (4.0) & & &  & (8.0) & (7.0) & (3.0) & (7.0) & (5.0) \\
\cline{1-7}
\cline{9-15}
\multirow{2}{2.1cm}{\centering\textbf{Rand-S}}
             & \multirow{2}{1.0cm}{\centering\textbf{fixed}} & $160.2$ & $178.3$ & $145.7$ & $175.6$ & $157.1$ & & \multirow{2}{2.1cm}{\centering\textbf{Rand-M}}
             & \multirow{2}{1.0cm}{\centering\textbf{fixed}} & 128.8 & 134.9 & 115.9 & 131.4 & 126.0 \\
              &  & (5.0) & (4.0) & (4.0) & (4.0) & (4.0) & & &  & (8.0) & (7.0) & (3.0) & (7.0) & (5.0) \\
\cline{1-7}
\cline{9-15}
\multirow{2}{2.1cm}{\centering\textbf{Max-Azim-S}}
             & \multirow{2}{1.0cm}{\centering\textbf{fixed}} & $156.3$ & $171.4$ & $139.9$ & $169.4$ & $150.3$ & & \multirow{2}{2.1cm}{\centering\textbf{Max-Azim-M}}
             & \multirow{2}{1.0cm}{\centering\textbf{fixed}} & 123.5 & 131.2 & 116.3 & 131.6 & 126.4 \\
              &  & (5.0) & (4.0) & (4.0) & (4.0) & (4.0) & & &  & (8.0) & (7.0) & (3.0) & (7.0) & (5.0) \\
\cline{1-7}
\cline{9-15}
\multirow{2}{2.1cm}{\centering\textbf{Oracle-S}}
             & \multirow{2}{1.0cm}{\centering\textbf{fixed}} & $103.4$ & $108.9$ & $106.5$ & $108.5$ & $105.4$ & & \multirow{2}{2.1cm}{\centering\textbf{Oracle-M}}
             & \multirow{2}{1.0cm}{\centering\textbf{fixed}} & 98.6 & 102.4 & 90.2 & 101.6 & 92.6 \\
              &  & (5.0) & (4.0) & (4.0) & (4.0) & (4.0) & & &  & (8.0) & (7.0) & (3.0) & (7.0) & (5.0) \\
\cline{1-7}
\cline{9-15}
\end{tabular}
}
\end{center}
\caption{Reconstruction error (mm/joint) for Pose-DRL and baselines on active-sequences on the selected Panoptic test splits. Results are shown both for the setting where the agent decides the number of views (auto) \emph{and} when using a fixed number of views. In the latter case, the number of views is set to the closest integer corresponding to the average in auto mode, rounded up. The baselines are also evaluated at this preset number of views. The average number of views are shown in parentheses. Pose-DRL models which automatically select the number of views outperform the heuristic baselines and fixed Pose-DRL models on all data splits, despite fusing estimates from fewer views on average. Left: Single-target mode (S), using DMHS as pose estimator. The agent significantly outperforms the baselines (e.g. 35  mm/joint improvement over \emph{Max-Azim} on multi-people data \emph{Maf} + \emph{Ult}). Right: Multi-target mode (M), using MubyNet as pose estimator. MubyNet is a more recent and accurate estimator, so the average errors are typically lower than the DMHS-counterparts. Automatic termination is useful in the multi-target setting as well, although it does not provide as drastic gains as in the single-target setup.}\label{t:main-results}
\end{table*}

\subsection{State-Action Representation}\label{sec:states}
To simplify notation, we here describe how the agent operates in a given time-freeze, and in this context will use $t$ to index actions within the active-view, as opposed to temporal structures. The state at step $t$ is the tuple $\pols^t = (\bB^t, \bX^t, \bC^t, \bu^t)$. Here $\bB^t \in \mathbb{R}^{H \times W \times C}$ is a deep feature map associated with the underlying 3d pose estimation architecture. $\bX^t = \{\bx_{t}, \tilde{\bx}, \bx_\star^{hist}\}$ where $\bx_{t}$ is the current individual pose estimate, $\tilde{\bx} = f(\bx_{1}, \dots, \bx_{t})$ is the current partially fused pose estimate, and $\bx_\star^{hist}$ is a history of fused predictions from 4 previous active-views. The matrix $\bC^t \in \mathbb{N}^{w \times h \times 2}$ consists of an \emph{angle canvas}, a discretized encoding\footnote{The camera sphere is discretized into $w$ bins in the azimuth direction and $h$ bins in the elevation direction, appropriately wrapped to account for periodicity. We set $w=9$ and $h=5$.} of the previously visited regions on the camera rig, as well as a similar encoding of the camera distribution over the rig. Finally, $\bu^t \in \mathbb{R}^2$ is an auxiliary vector holding the number of actions taken and the number of people detected. 
 
For action selection we use a deep stochastic policy $\pith(\pola^t|\pols^t)$ parametrized by $\bs{w}$ which predicts the action $A^t = \{\theta^t_a, \theta^t_e, c^t\}$. Here $(\theta^t_a, \theta^t_e)$ is the azimuth-elevation angle pair, jointly referred to as \emph{viewpoint selection}, and $c^t$ is a Bernoulli variable indicating whether to continue to the next active-view (occurring if $c^t=1$), referred to as the \emph{continue} action. To produce action probabilities the base feature map $\bB^t$ is fed through two convolutional blocks which are shared between the viewpoint selection and continue actions. The output of the second convolutional block is then concatenated with $\bs{X}^t$, $\bs{C}^t$ and $\bu^t$ and fed to viewpoint selection- and continue-branches with individual parameters. Both action branches consist of 3 fully-connected layers with $\tanh$ activations. The probability of issuing the continue action is computed using a sigmoid layer:
\begin{equation}\label{eq:policy_done}
\pith(c^t = 1 | \pols^t) = \sigm{\bw_{c}^\top \bz^t_c + b_c}
\end{equation}
where $\bw_c$ and $b_c$ are trainable weights and bias, and $\bz^t_c$ is the output from the penultimate fully-connected layer of the \emph{continue} action branch. 

Due to the periodic nature of the viewpoint prediction task we rely on von Mises distributions for sampling the spherical angles. We use individual distributions for the azimuth and elevation angles. The probability density function for the azimuth is given by:
\begin{equation}\label{eq:von_mises_azim}
\begin{split}
    &\piw\left(\theta^t_a | \pols^t\right) \\
    &= \frac{1}{2 \pi I_0(m_a)} \exp\{m_a \cos(\theta^t_a - \tilde{\theta}_a(\bw_a^\top \bz^t_a + b_a))\}
\end{split}
\end{equation}
where $I_0$ is the zeroth-order Bessel function, normalizing \eqref{eq:von_mises_azim} to a proper probability distribution over the unit circle $[-\pi,\pi]$. Here $\tilde{\theta}^a$ is the mean of the distribution (parametrized by the neural network), $m_a$ is the precision parameter,\footnote{We treat the precision parameters as constants but increase them over training to focus the policy on high-reward viewpoints.} $\bw_a$ and $b_a$ are trainable weights and bias, respectively, and $\bz^t_a$ comes from the penultimate fully-connected layer of the viewpoint selection action branch. The support for the azimuth angle should be on a full circle $[-\pi, \pi]$, and hence we set
\begin{equation}\label{eq:azim_mean_tanh}
    \tilde{\theta}_a(\bw_a^\top \bz^t_a + b_a) = \pi \tanh(\bw_a^\top \bz^t_a + b_a)
\end{equation}
The probability density function for the elevation angle has the same form \eqref{eq:von_mises_azim} as that for the azimuth. However, as seen in \Figure{fig:overview-one-mc}, the range of elevation angles is more limited than for the azimuth angles. We denote this range $[-\kappa, \kappa]$ and the mean elevation angle thus becomes\footnote{With notation analogous to that of the azimuth angle, cf. \eqref{eq:azim_mean_tanh}.}
\begin{equation}\label{eq:elev_mean_tanh}
    \tilde{\theta}_e(\bw_e^\top \bz^t_e + b_e) = \kappa \tanh(\bw_e^\top \bz_e^t + b_e)
\end{equation}
In practice, when sampling elevation angles from the von Mises, we reject samples outside the range $[-\kappa, \kappa]$.
\begin{figure*}[!htbp]
  \begin{center}
  \begin{subfigure}[b]{0.285\textwidth}
    \includegraphics[width=\textwidth]{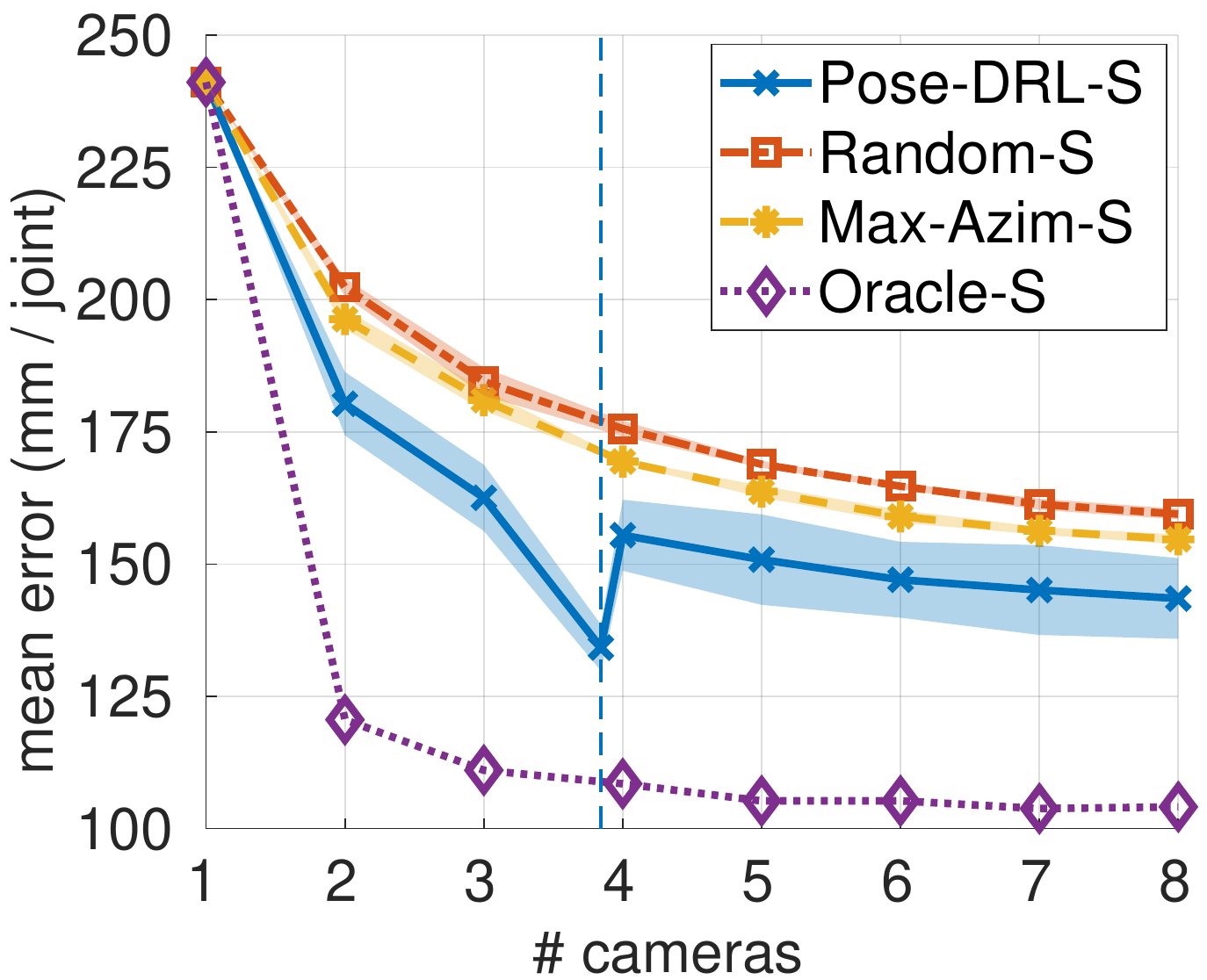}
  \end{subfigure}
  \hfill
  \begin{subfigure}[b]{0.285\textwidth}
    \includegraphics[width=\textwidth]{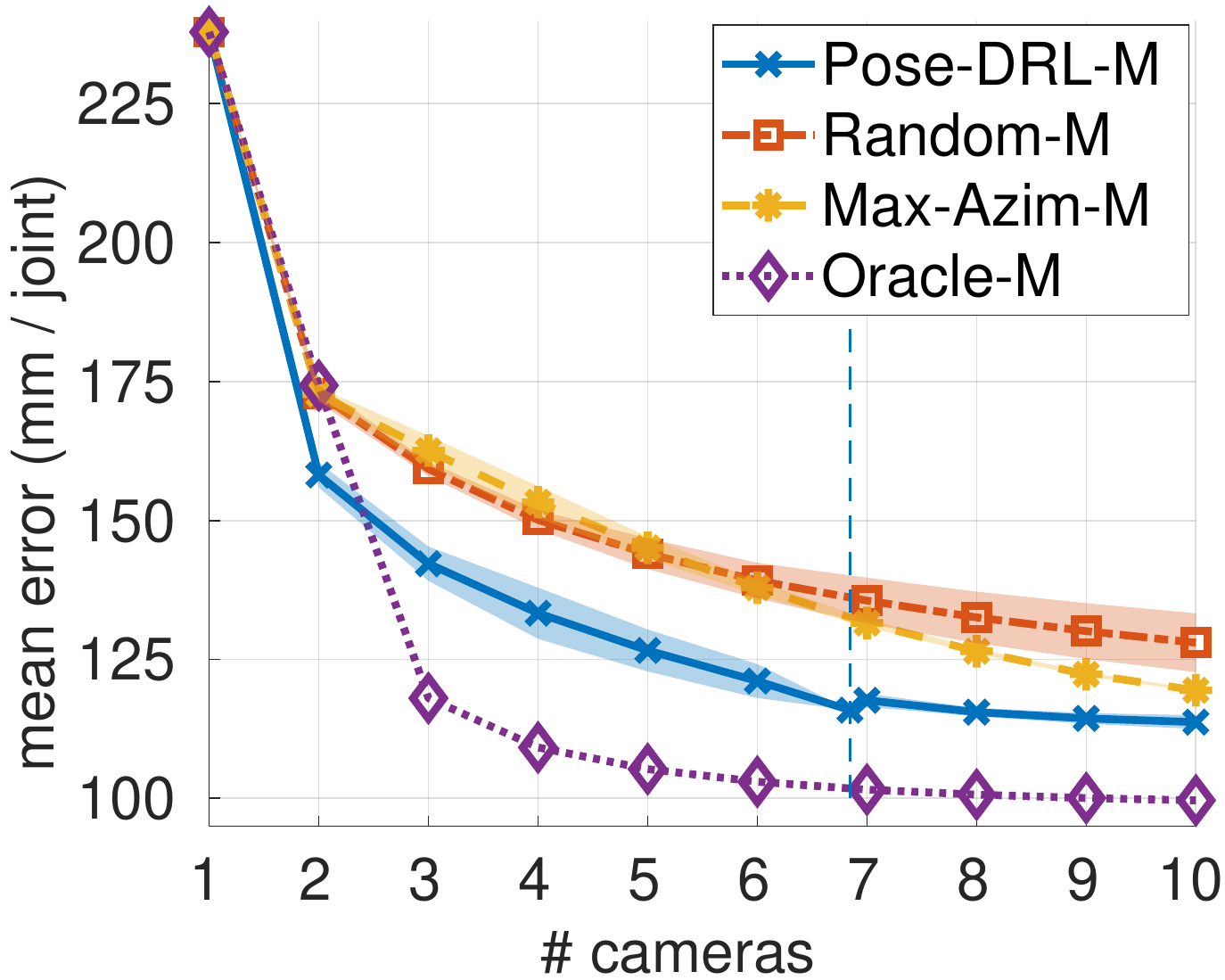}
  \end{subfigure}
  \hfill
  \begin{subfigure}[b]{0.285\textwidth}
    \includegraphics[width=\textwidth]{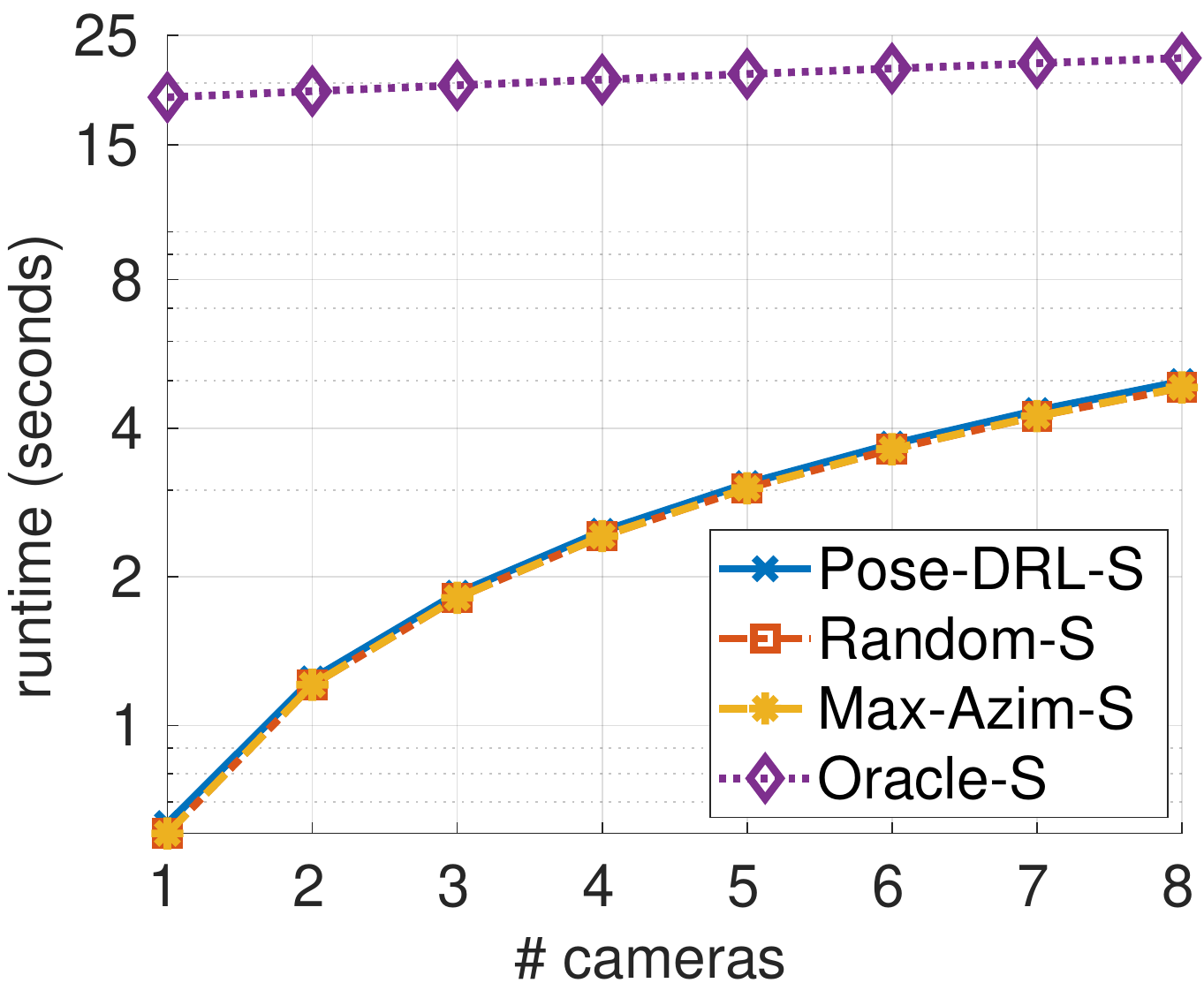}
  \end{subfigure}
\caption{How the number of views affects pose estimation error and runtimes of Pose-DRL and baselines on multi-people data (union of \emph{Mafia} and \emph{Ultimatum} test sets). We show mean and 95\% confidence intervals over 5 seeds. Left: Reconstructing a single target person. Estimation error reduces with added viewpoints, and the agent consistently outperforms the non-oracle baselines. The automatic \emph{continue} action (dashed line at 3.8 views on average) yields significantly lower reconstruction errors than any fixed viewpoint schemes. Hence the auto-model clearly provides the best speed-accuracy trade-off. Middle: Simultaneously reconstructing all persons. The agent outperforms the heuristic baselines in this setting too. Adaptively determining when to continue to the next active-view (6.8 views on average) yields better results than fusing from 7 cameras all the time. The gain is not as pronounced as in the single-target case, since more viewpoints typically leads to increased estimation accuracy for some of the persons. Right: Runtime of the Pose-DRL agent and baselines vs. number of views (log scale). The oracle always needs to evaluate the deep pose estimation system and detector for all cameras due to its need to sort from best to worst, independently of the number of viewpoints, which explains its high runtime. Our agent is as fast as the heuristic baselines.}\label{fig:err-vs-fix}
 \end{center}
\end{figure*}

\subsection{Reward Signal for Policy Gradient Objective}\label{sec:reward}
The agent should strike a balance between choosing sufficiently many cameras so the resulting 3d pose estimate is as accurate as possible, while ensuring that not too many cameras are visited, to save processing time. As described earlier, the two types of actions are \emph{viewpoint selection} and \emph{continue}. We will next cover the reward functions for them. \\ \\
\noindent\textbf{Viewpoint selection reward.} At the end of an active-view we give a reward which is inversely proportional to the ratio between the final and initial reconstruction errors within the active-view. We also give a penalty $\epsilon=2.5$ each time the agent goes to an already visited viewpoint. Thus the viewpoint selection reward is:
\begin{equation}\label{eq:reward_view}
r^t_v = \begin{cases}
0, & \text{if } c^t=0 \text{ and view not visited} \\
-\epsilon, & \text{if } c^t=0 \text{ and view visited before} \\
1 - \frac{\varepsilon^{k}}{\varepsilon^1}, & \text{if } c^t=1
\end{cases}
\end{equation}
where $k$ is the number of views visited prior to the agent issuing the \emph{continue} action ($c^t = 1$), $\varepsilon^1$ is the reconstruction error associated with the initial viewpoint, and $\varepsilon^k$ denotes the final reconstruction error, i.e. $\varepsilon^k = \|\bx_\star - \bx_{\text{gt}}\|_2^2$. Here $\bx_\star$ is the final fused pose estimate for the active-view, cf. \eqref{eq:pose_fusion}, and $\bx_{\text{gt}}$ is the ground-truth 3d pose for the time-freeze.  \\ \\
\noindent\textbf{Continue action reward.} The \emph{continue} action has two purposes: \emph{(i)} ensure that not too many viewpoints are visited to reduce computation time, and \emph{(ii)} stop before suboptimal viewpoints are explored, which could happen if the agent is forced to visit a preset number of viewpoints. Therefore, the \emph{continue} action reward is:
\begin{equation}\label{eq:reward_cont}
r^t_c = \begin{cases}
1 - \frac{\min\limits_{j \in \{t+1,\dots,k\}}\varepsilon^j}{\varepsilon^t} - \tau, & \text{if } c^t=0 \\
1 - \frac{\varepsilon^{k}}{\varepsilon^1}, & \text{if } c^t=1
\end{cases}
\end{equation}
At each step that the agent decides \emph{not} to continue to the next active-view ($c^t=0$), the agent is rewarded relative to the ratio between the error at the best future stopping point within the active-view (with lowest reconstruction error) and the error at the current step. If in the future the agent selects viewpoints that yield lower reconstruction error the agent is rewarded, and vice verse if the best future error is higher. In addition, the agent gets a penalty $\tau$ at each step, which acts as an \emph{improvement threshold}, causing the reward to become negative unless the ratio is above the specified threshold $\tau$. This encourages the agent not to visit many viewpoints in the current active-view unless the improvement is above the given threshold. On the validation set we found $\tau = 0.07$ to provide a good balance.\\ \\
\noindent\textbf{Policy gradient objective.} We train the Pose-DRL network in a policy gradient framework, maximizing expected cumulative reward on the training set with objective
\begin{equation}\label{eq:pol_obj}
J(\bw) = \mathbb{E}_{\bss \sim \pith} \left[\sum_{t=1}^{\left|\bss\right|} r^t\right]
\end{equation}
where $\bss$ denotes state-action trajectories, and the reward signal $r^t=r_v^t + r_c^t$, cf. \eqref{eq:reward_view} - \eqref{eq:reward_cont}. We approximate the gradient of the objective \eqref{eq:pol_obj} using REINFORCE \cite{Williams-ml-1992}. 

\subsection{Active Pose Estimation of Multiple People}\label{sec:multiple-people}
So far we have explained the Pose-DRL system that estimates the pose of a target person, assuming it is equipped with a detection-based single person estimator. This system can in principle estimate multiple people by generating active-sequences for each person individually. However, to find a \emph{single} active-sequence that reconstructs \emph{all} persons, one can equip Pose-DRL with an image-level multi-people estimator instead. In that case, the state representation is modified to use the image level feature blob from the multi-people estimator ($\bs{B}_t$ in \Figure{fig:overview-one-mc}). The reward signal used when learning to reconstruct all people is identical to \eqref{eq:reward_view} - \eqref{eq:reward_cont}, except that the rewards are averaged over the individual pose estimates. Thus Pose-DRL is very adaptable in that the underlying pose estimator can easily be changed.

\section{Experiments}\label{sec:results}
\noindent{\bf Dataset.} We use diverse scenes for demonstrating and comparing our active pose estimation system, considering complex scenes with multiple people (\emph{Mafia}, \emph{Ultimatum}) as well as single person ones (\emph{Pose}). The motions range from basic poses to various social games. Panoptic provides data as 30 FPS-videos which we sample to 2 FPS, making the data more manageable in size. It also increases the change in pose between consecutive frames.

The data we use consists of the same 20 scenes as in \cite{NIPS2019_8646}. The scenes are randomly split into training, validation and test sets with 10, 4 and 6 scenes, respectively. Since we split the data over the scenes, the agent needs to learn a general look-around-policy which adapts to various circumstances (scenarios and people differ between scenes). All model selection is performed exclusively on the training and validation sets; final evaluations are performed on the test set. The data consists of 343k images, of which 140k are single-person and 203k are multi-people scenes. \\ \\
\noindent{\bf Implementation details.} We attach Pose-DRL on top of the DMHS monocular pose estimation system \cite{popa2017deep}. In the multi-people setting described in \Section{sec:multiple-people} we instead use MubyNet \cite{zanfir2018deep}. Both estimators were trained on Human3.6M \cite{h36m_pami}. To avoid overfitting we do not to fine-tune these on Panoptic, and instead emphasize how Pose-DRL can select good views with respect to the underlying estimation system (but joint training is possible). We use an \emph{identical set of hyperparameters} when using DMHS and MubyNet, except the improvement threshold $\tau$, which is $-0.07$ for DMHS and $-0.04$ for MubyNet, which shows that Pose-DRL is robust with respect to the pose estimator used. We use median averaging for fusing poses, cf. \eqref{eq:pose_fusion}. \\ \\
\noindent{\bf Training.} We use 5 active-sequences, each consisting of length 10, to approximate the policy gradient, and update the policy parameters using Adam \cite{kingma2014adam}. As standard, to reduce variance we normalize cumulative rewards for each episode to zero mean and unit variance over the batch. The maximum trajectory length is set to 8 views including the initial one (10 in the multi-target mode, as it may require more views to reconstruct all people).
The \emph{viewpoint selection} and \emph{continue} actions are trained jointly for 80k episodes. The learning rate is initially set to \verb|5e-7| and is halved at 720k and 1440k agent steps. We linearly increase the precision parameters $m_a$ and $m_e$ of the von Mises distributions from $(1,10)$ to $(25,50)$ in training, making the viewpoint selection increasingly focused on high-rewarding regions as training proceeds.\\ \\
\noindent{\bf Baselines.} To evaluate our active human pose estimation system we compare it to several baselines, similar to \cite{NIPS2019_8646}. For fair comparisons, the baselines use the same pose estimator, detector and matching approach. All methods obtain the same initial random view as the agent at the start of the active-sequence. We design the following baselines: i) \emph{Random:} Selects $k$ different random views; ii) {\it Max-Azim:} Selects $k$ different views equidistantly with respect to the azimuth angle. At each azimuth angle it selects a random elevation angle; iii) \emph{Oracle:} Selects as next viewpoint the one that minimizes the fused 3d pose reconstruction when combined with pose estimates from all viewpoints observed so far (averaged over all people in the multi-target setting). This baseline cheats by extensively using ground-truth information, and thus it shown as a lower bound with respect to reconstruction error. In addition to cheating during viewpoint selection, the oracle is also impractically slow  since it requires computing pose estimates for \emph{all} available viewpoints and exhaustively computing errors for all cameras in each step.

\begin{figure}[!htbp]
\begin{center}
\includegraphics[width=.38\textwidth]{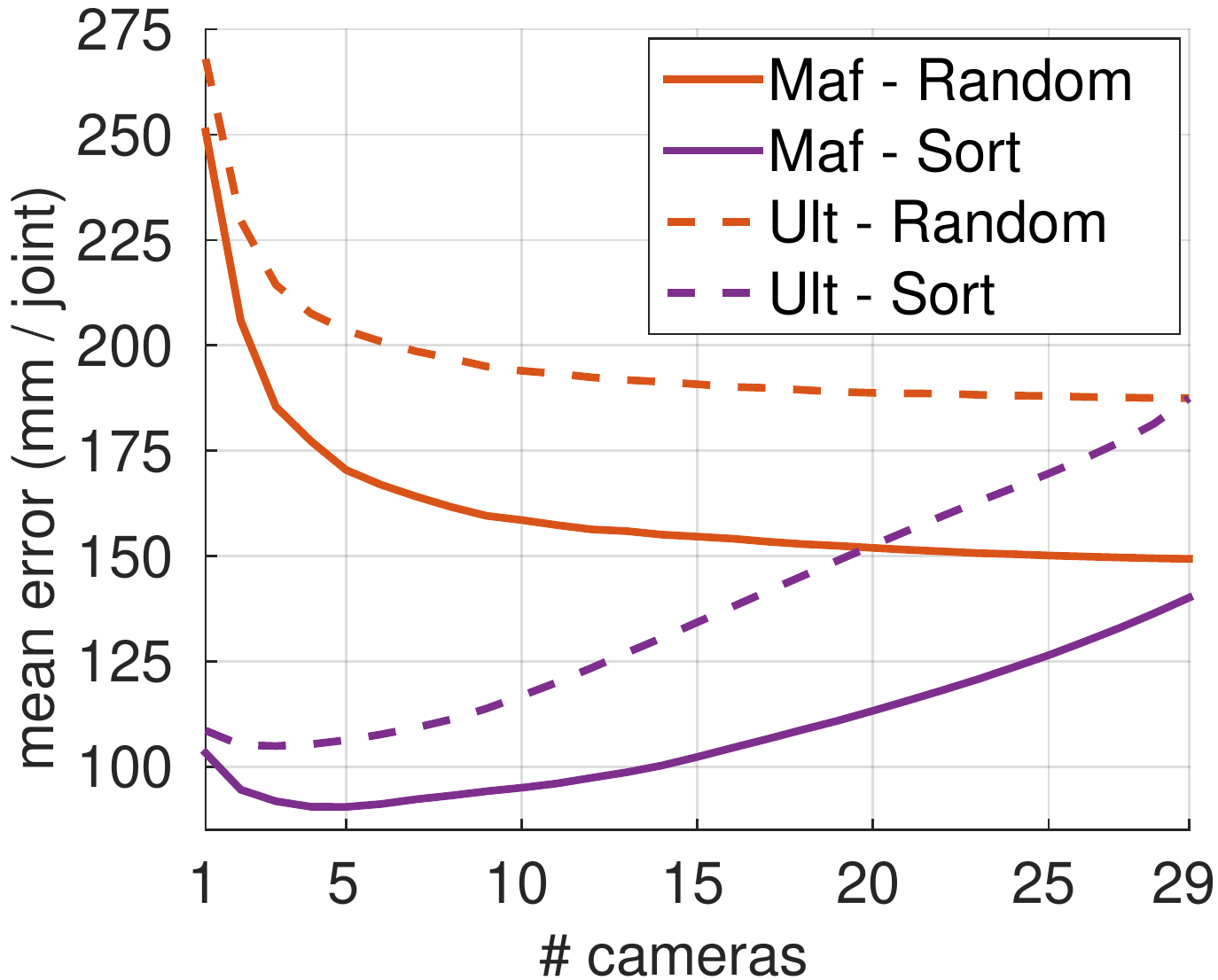}
\caption{Per-joint pose reconstruction error for the monocular human pose estimation architecture DMHS vs. number of viewpoints, both when randomly choosing viewpoints, and when using an sorting strategy which selects viewpoints in ascending order of individual reconstruction error (note that this requires ground-truth). Results shown for multi-people data (\emph{Mafia}, \emph{Ultimatum}) on the CMU Panoptic dataset. For a good viewpoint selection policy such as \emph{Sort}, estimation accuracy only improves when adding a few extra cameras, but then begins to deteriorate, indicating the need to adaptively terminate viewpoint selection early enough.}\label{fig:err-all-cams}
\end{center}
\end{figure}
\begin{figure*}[!htbp]
    \begin{center}
    \includegraphics[width=0.93\textwidth]{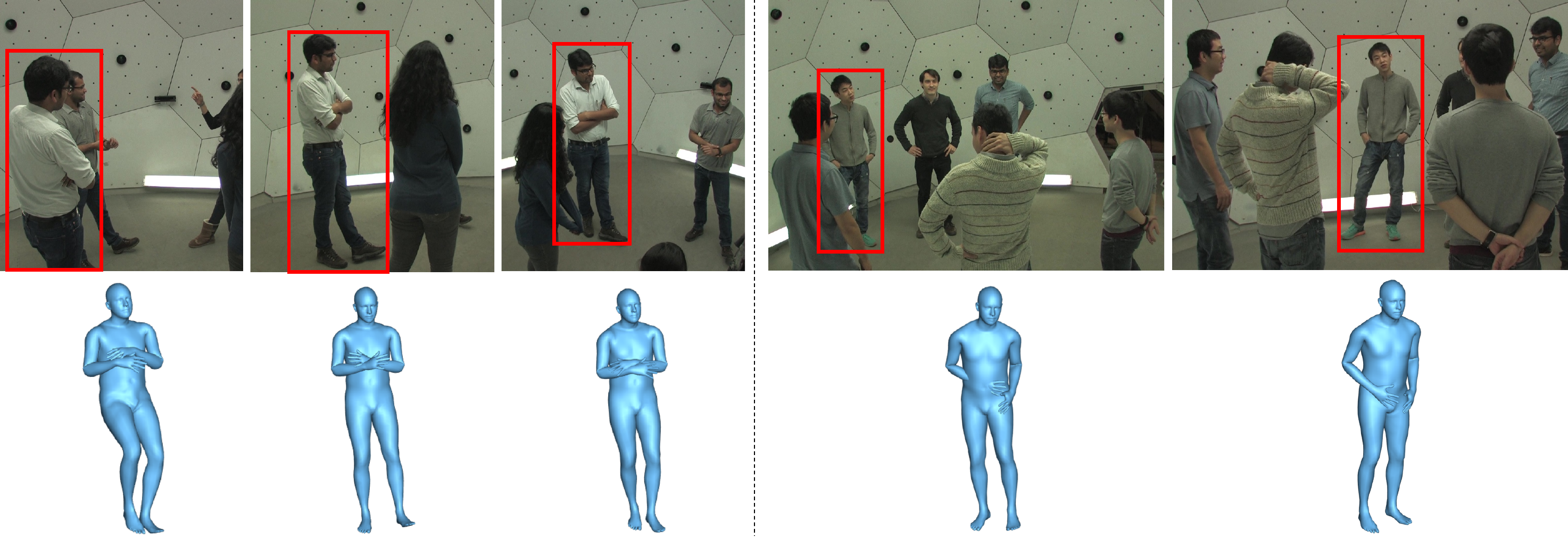}
    \caption{Visualizations of Pose-DRL reconstructing a given target person (red bounding box). Left: A \emph{Mafia} test scene. The target is viewed from behind and is partially visible in the first view, producing the poor first estimate. As the agent moves to the next view, the person becomes more clearly visible, significantly improving the estimate. The last view from the front further increases accuracy. The agent decides to terminate after three views with error decreasing from 200.1 to 120.9 mm/joint. Right: An \emph{Ultimatum} test scene where the agent only requires two viewpoints prior to automatically continuing to the next active-view. The target person is only partially visible in the initial viewpoint, and the right arm that is not visible results in a non-plausible configuration in the associated estimate. As the agent moves to the next viewpoint the person becomes fully visible, and the final fused estimate is both physically plausible and accurate. The reconstruction error reduces from 160 to 104 mm/joint.}\label{fig:showcase-single}
    \end{center}
\end{figure*}
\subsection{Quantitative Results}\label{sec:main-results}
We report results both for the Pose-DRL agent that tracks and reconstructs a single target person (possibly in crowded scenes) and for the Pose-DRL model which actively estimates poses for all persons in the scene, cf. \Section{sec:multiple-people}. 
Pose-DRL is trained over 5 different random initializations of the policy network, and we report average results. In each case, training the model 80k steps gave best results on the validation set, so we use that. Also, for the heuristic baselines we report average results over 5 seeds (the oracle is deterministic).

\begin{figure*}[!htbp]
    \begin{center}
    \includegraphics[width=0.99\textwidth]{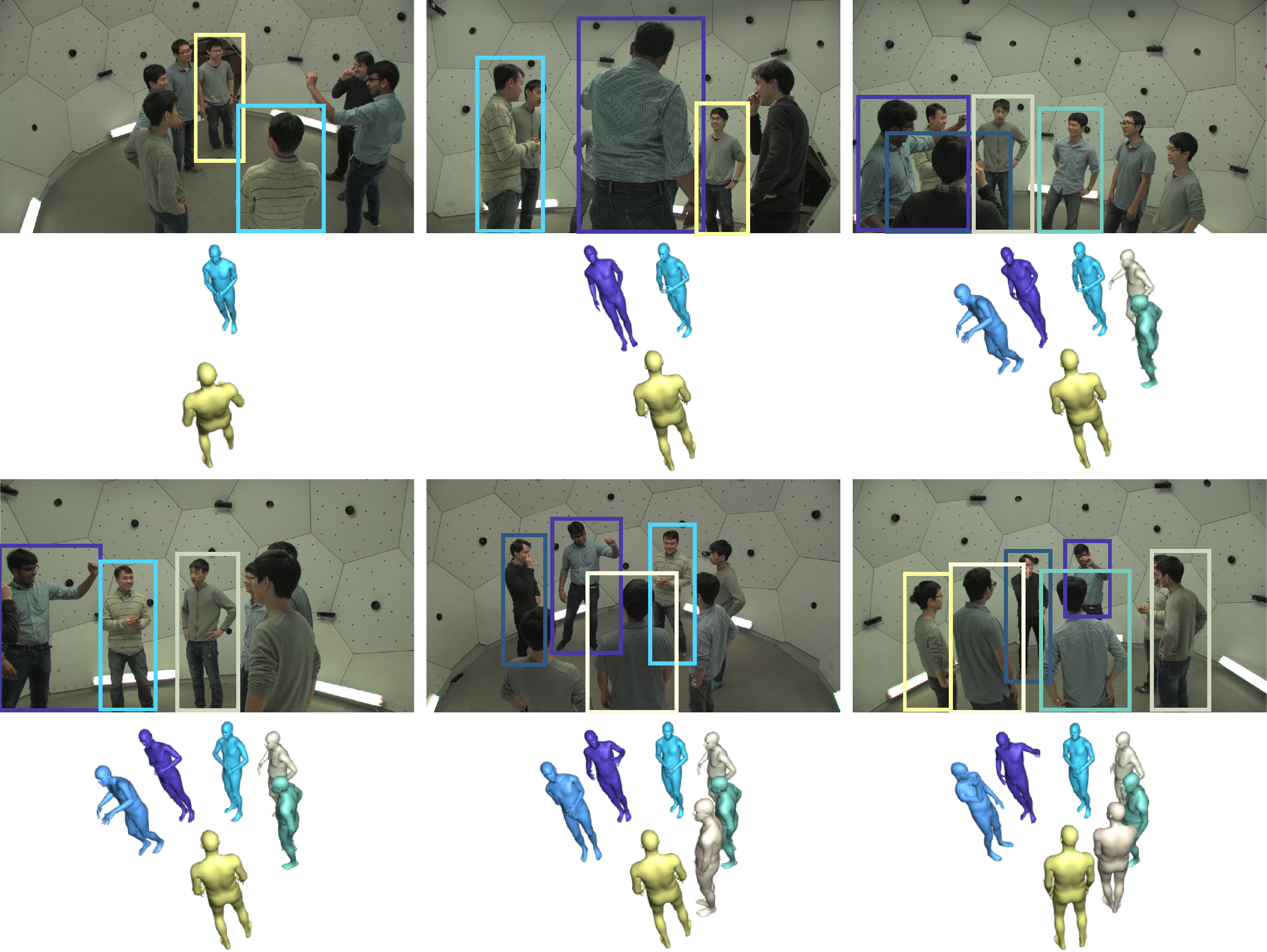}
    \caption{Visualization of how Pose-DRL performs multi-target pose estimation for an \emph{Ultimatum} test scene. In this example the agent sees six viewpoints prior to automatically continuing to the next active-view. The mean error decreases from 358.9 to 114.6 mm/joint. Only two people are detected in the initial viewpoint, but the number of people detected increases as the agent inspects more views. Also, the estimates of already detected people improve as they get fused from multiple viewpoints. }\label{fig:showcase-multi-1}
    \end{center}
\end{figure*}

\begin{figure*}[!htbp]
    \begin{center}
    \includegraphics[width=0.99\textwidth]{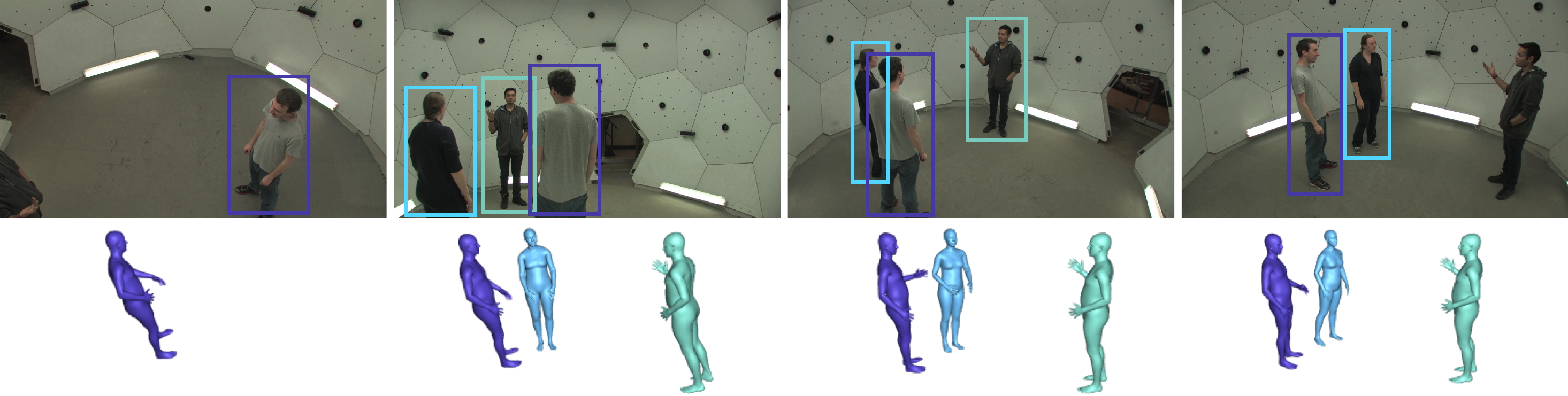}
    \caption{Visualization of how Pose-DRL performs multi-target pose estimation an \emph{Ultimatum} validation scene. The agent chooses four viewpoints prior to automatically continuing to the next active-view. The mean error decreases from 334.8 to 100.9 mm/joint. Only one of the persons is visible in the initial viewpoint, and from a poor angle. This produces the first, incorrectly tilted pose estimate, but the estimate improves as the agent inspects more viewpoints. The two remaining people are successfully reconstructed in subsequent viewpoints.}\label{fig:showcase-multi-2}
    \end{center}
\end{figure*}

Our agent is compared to the baselines on the Panoptic test set on active-sequences consisting of 10 active-views.
\Table{t:main-results} presents reconstruction errors. \Figure{fig:err-vs-fix} shows how the the number of selected views affects accuracy and runtimes. For visualizations\footnote{We use SMPL \cite{SMPL2015} for the 3d shape models.} of Pose-DRL, see Fig. \ref{fig:showcase-single} - \ref{fig:showcase-multi-2}. \\ \\
\noindent\textbf{Single-target estimation.} It is clear from \Table{t:main-results} (left) and \Figure{fig:err-vs-fix} (left) that Pose-DRL outperforms the heuristic baselines, which is particularly pronounced for multi-people data. In such scenes, the view selection process is more delicate, as it requires avoiding cameras where the target is occluded. We note that the automatically stopping agent yields by far the most accurate estimates, which shows that it is capable of continuing to the next active-view when it is likely that the current one does not provide any more good views. Thus it is often better to fuse a few accurate estimates than including a larger set of poorer ones.\\ \\
\noindent\textbf{Multi-target estimation.} From \Table{t:main-results} (right) and \Figure{fig:err-vs-fix} (middle) we see that the agent outperforms the heuristic baselines as in the case with a single target. Automatic view selection termination does not yield as big improvements in accuracy as in the single-target case. In the single-target setting the agent stops early to avoid occluded and bad views, but when reconstructing all people there is more reason to keep selecting additional views to find some views which provide reasonable estimates for each person. This also explains the decreased gaps between the various methods -- there may be many sets of cameras which together provide a fairly similar result when averaged over all people in the scene. A future improvement could include selective fusing a subset of estimates in each view. Running in auto mode still yields more accurate estimates than fixed schemes which use a larger number of views.\\ \\
\noindent\textbf{Runtimes.} The runtimes\footnote{Shown for DMHS-based systems. Using MubyNet (which requires 1.01 seconds per image) gives runtime curves which look qualitatively similar.} for Pose-DRL and baselines are shown in
\Figure{fig:err-vs-fix}. DMHS and Faster R-CNN require $0.50$ and $0.11$ seconds per viewpoint, respectively, which constitutes the bulk of the processing time. The policy network has an overhead of about 0.01 seconds per action, negligible in relation to the pose estimation system.

\begin{table}[!htbp]
\begin{center}
\scalebox{0.95}{
\begin{tabular}{|c|c|c|c|c|}
\hline
\textbf{\Centerstack{Model}} & \textbf{\Centerstack{Settings}} & \textbf{\Centerstack{Maf}} & \textbf{Ult} & \textbf{Pose} \\
\hline
\multirow{3}{1cm}{\centering\textbf{Pose-DRL}} 
             & \textbf{full model} & $144.7$ (5) & $157.5$ (4) & $135.1$ (4) \\
\cline{2-5}
              & $\bB^t$ \textbf{only} & $153.5$ (5) & $166.9$ (4) & $134.4$ (4) \\
\cline{2-5}
             & \textbf{reset} & $152.5$ (5) & $160.8$ (4) & $133.4$ (4) \\
\hline
\end{tabular}
}
\end{center}
\caption{Ablations on the test sets, showing the effect of removing certain components of the DMHS-based Pose-DRL system. Results (errors, mm/joint) are for models that select a fixed number of views (shown in parentheses), where the number of views are the same as in \Table{t:main-results}. Providing more information than the base feature map $\bB^t$ is crucial for crowded scenes with multiple people (\emph{Maf}, \emph{Ult}), as is including previous pose estimates in the current pose fusion.}\label{t:ablations}
\end{table}

\subsection{Ablation Studies}\label{sec:ablations}
In this section we compare the full agent to versions lacking parts of the model: i) providing only the base feature map $\bB^t$, and ii) not propagating the fused reconstruction $\bx^{t}_{\star}$ to the next active-view (\emph{reset}), cf. \eqref{eq:pose_fusion}. The results are given in \Table{t:ablations}, and show that the full model outperforms the stripped-down versions for multi-people data (\emph{Mafia}, \emph{Ultimatum}), while simpler single-people data in \emph{Pose} is not sensitive to removing some parts of the model. There is significantly more room for intelligent decision making for complex multi-people data, where the model has to avoid occlusions, and thus it requires a stronger state description and fusion approach. In contrast, selecting views in single-people scenes is less fragile to the particular camera choices as there is no risk of choosing views where the target is occluded.

\section{Conclusions}
In this paper we have presented \emph{Pose-DRL}, a fully trainable deep reinforcement-learning based active vision model for human pose estimation. The agent has the freedom to move and explore the scene spatially and temporally, by selecting informative views that improve its accuracy. The model learns automatic stopping conditions for each moment in time, and transition functions to the next temporal processing step in video. We showed in extensive experiments -- designed around the dense Panoptic multi-camera setup, and for complex scenes with multiple people -- that Pose-DRL produces accurate estimates, and that our agent is robust with respect to the underlying pose estimator used. Moreover, the results show that our model learns to select an adaptively selected number of informative views which result in considerably more accurate pose estimates compared to strong multi-view baselines.
 
Practical developments of our methodology would include e.g. real-time intelligent processing of multi-camera
video feeds or controlling a drone observer. In the latter case the model would further benefit from being extended to account for physical constraints, e.g. a single camera and limited speed. Our paper is a key step since it presents fundamental methodology required for future applied research.\\ \\
{\small \noindent{\bf Acknowledgments:} This work was supported by the European Research Council Consolidator grant
SEED, CNCS-UEFISCDI PN-III-P4-ID-PCE-2016-0535 and PCCF-2016-0180, the
EU Horizon 2020 Grant DE-ENIGMA, Swedish Foundation for Strategic Research (SSF) Smart Systems Program, as well as the Wallenberg AI, Autonomous Systems and Software Program (WASP) funded by the Knut and Alice Wallenberg Foundation. Finally, we would like to thank Alin Popa, Andrei Zanfir, Mihai Zanfir and Elisabeta Oneata for helpful discussions and support.}

{\small\textbf{}
\bibliographystyle{aaai}
\bibliography{bibfile}
}

\clearpage


\section*{Supplementary material}

# Deep Reinforcement Learning for Active Human Pose Estimation
## Supplementary Material

Erik Gärtner[1][*], Aleksis Pirinen[1][*], Cristian Sminchisescu[1,2]

[1]Department of Mathematics, Faculty of Engineering, Lund University
[2]Google Research
{erik.gartner, aleksis.pirinen, cristian.sminchisescu}@math.lth.se

In this supplemental we provide additional insights into our Pose-DRL model. Details of the network architecture are provided in § 1. Further model insights and dataset details are provided in § 2. A description of how we handle missed detections or failed matchings are given in § 3. Finally, additional visualizations are shown in § 4.

## 1 Model Architecture

See Fig. 1 for a description of the Pose-DRL architecture. The underlying pose estimation networks, DMHS (Popa, Zanfir, and Sminchisescu 2017) and MubyNet (Zanfir et al. 2018), as well as our agent were implemented in Caffe (Jia et al. 2014) and MATLAB. For the Faster R-CNN detector (Ren et al. 2015) we used a publicly available Tensorflow (Abadi et al. 2016) implementation,[1] with ResNet-101 (He et al. 2016) as base feature extractor.

|  | 10% best | 10% worst | Rest |
|---|---|---|---|
| **Mafia** | 52 % | 2% | 46% |
| **Ultimatum** | 67% | 1% | 32% |
| **Pose** | 24% | 2% | 74% |
| **All** | 43% | 2% | 55% |

Table 1: Pose-DRL agent's selection statistics of good / bad viewpoints on the test set splits. The agent consistently chooses a high percentage of good cameras while avoiding bad cameras. Note that randomly choosing cameras would result in always having 10% chosen among the 10% best cameras, and similar for the 10% worst cameras.

## 2 Additional Insights and Details

**More about runtimes.** All experiments reported in this supplementary material and in the main paper were performed using an Ubuntu workstation using a single Titan V100. Training the Pose-DRL policy from scratch took about 70 hours after having pre-computed all DMHS /

[1]https://github.com/smallcorgi/Faster-RCNN_TF

|  | Train | Val | Test | All |
|---|---|---|---|---|
| **Mafia** | 53,100 | 27,900 | 33,728 | 114,728 |
| **Ultimatum** | 27,960 | 4,340 | 55,825 | 88,125 |
| **Pose** | 51,079 | 29,672 | 59,288 | 140,039 |
| **All** | 132,139 | 61,912 | 148,841 | 342,892 |

Table 2: Breakdown of the subset of the Panoptic dataset used in this work for the training, validation and test splits, respectively. In each cell is shown the number of images. The fourth row shows the total number of images in the train, val and test splits (summed over *Mafia*, *Ultimatum* and *Pose*). The fourth column shows the total number of images for *Mafia*, *Ultimatum* and *Pose* (summed over the train, val and test splits). The bottom-right cell shows the total number of images in the entire used dataset.

MubyNet features, Faster R-CNN bounding boxes and instance features. When presenting the runtimes (see Figure 2 in the main paper) we include the time needed to compute these detections and features.

**Quality of selected viewpoints.** To obtain further insights into which cameras the agent is selecting on average, we tracked how often the agent selects good vs bad viewpoints (for the DMHS-based model). Specifically, for each selected camera in the various test set splits, we sorted it into being in either the 10% best or worst cameras based on associated individual reconstruction error. The results are shown in Table 1. It can be seen that the agent typically selects among the best while avoiding the worst viewpoints. The viewpoint errors are more uniform for the single-people *Pose* scenes, since there are no viewpoints where the target is occluded, hence the camera selection statistics are also more uniform for *Pose*.

**Further dataset insights.** In Table 2 we show how we randomly split the Panoptic dataset (Joo et al. 2015) into train, test, and validation sets.

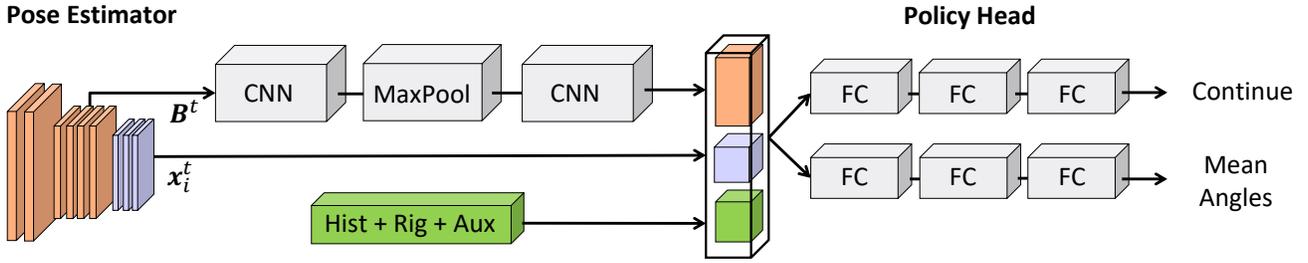

Figure 1: Pose-DRL network architecture. The pose estimator is shown to the left (we have shown results for two different pose estimators, DMHS and MubyNet, but any other moncoular pose estimator would work). When using the single person pose estimator DMHS, the input is a bounding box containing the target person, and its convolutional feature map $B_t$ forms the base state of the agent. For the multi-person estimator MubyNet, the full image is instead fed as input and the associated feature map $B_t$ is used as the base state. Next, $B_t$ is processed by two convolutional layers with ReLU-activations (first conv: $3 \times 3$ kernel, stride 1, output dimension $21 \times 21 \times 8$; max pool: $2 \times 2$ kernel, stride 2, output dimension $11 \times 11 \times 8$; second conv: $3 \times 3$ kernel, stride 1, output dimension $9 \times 9 \times 4$). It is then concatenated with the pose prediction information $x_i^t$ for the current active-view, a history of the last 4 fused pose estimates from previous active-views (*Hist*), camera rig information (*Rig*), containing both a description of the camera rig as well as the agent's current and previously visited viewpoints within the rig, and auxiliary information (*Aux*) with the number of actions taken and number of people detected. Note that pose information is used in the single-target mode only; for the multi-person setting we omit pose information in the state space as there may be a variable number of persons per scene. However, in this setting the agent instead has access to image level information. See more about the state space in §4.2 in the main paper. The concatenated state is subsequently fed to the two action branches: the *continue* action branch (top) and the *viewpoint selection* action branch (bottom). Both branches use tanh-activations for the hidden fully-connected (FC) layers. For the *continue* action branch, the output is turned into a *continue* probability through a sigmoid-layer, cf. (2) in the main paper. For the *viewpoint selection* action branch, the azimuth and elevation mean angles are produced by a scaled tanh-layer, cf. equations (4) and (5) in the main paper. In the *continue* action branch the three FC-layers have 512, 512, and 1 output neurons each respectively, while the *viewpoint selection* action branch's three FC-layers have 1024, 512, and 2 output neurons, respectively.

## 3 Handling Missed Detections or Matchings

For an overview of how we detect and match multiple people, refer to §3.2 in the main paper. In this section we describe what happens in case some persons are not detected or matched. For the detection-based DMHS-version of Pose-DRL, if in a viewpoint there are no detections, or if no detection has a matching cost below the threshold $\mathcal{C}$, the underlying pose estimator is computed on the entire input image to obtain a base state descriptor $B_t$ for decision making (no associated pose is fused in this case).

It is possible that one or several persons are not detected in a single viewpoint in an active-view. In this case the pose estimate is set to the fused estimate from the previous active-view as a backup. In case a previous estimate also does not exist (could happen e.g. in the initial active-view of an active-sequence), to be able to compute a reconstruction error we set a placeholder pose estimate where each joint is equal to the center hip location of the ground-truth. Naturally, this is an extremely poor and implausible estimate, but it is used only to be able to compute an error (another option would be to not include such an estimate when computing average errors, but that would not penalize the fact that a person was never detected and reconstructed).

## 4 Additional Visualizations of Pose-DRL

In Fig. 2 - 3 we show two additional visualizations of how Pose-DRL performs single-target pose estimation in active-views from the Panoptic (Joo et al. 2015) test set we have used in this work. We use SMPL (Loper et al. 2015) for the 3d shape models (both here and in the main paper), and use per-joint median averaging for fusing poses. As it is referenced in the visualizations, we show the equation for a partially fused pose (for the first $j$ steps) within an active-view[2] below:

$$\tilde{x} = f(x_1, \ldots, x_i) \quad (1)$$

### 4.1 Using Pose-DRL the Wild

The dense CMU Panoptic studio provides a powerful environment for training and evaluating our proposed model, however it is also interesting to test the model's applicability in the the real world. To this end we captured data with an off-the-shelf smartphone and used internal sensors to estimate the camera pose matrix for each image. This simple process of walking around subjects while they stand still emulates the *time-freeze* setup in Panoptic and allows us to test our model in the real world. Please note that neither the 3d pose estimation network nor the policy was re-trained; only the instance detector was refined to produce accurate appearance models for the detected people. See Fig. 4 for resulting visualizations. Please note we obtained consent from the people shown.

---
[2]For active-sequence processing, the agent also fuses temporally by adding the previous fused estimate; see eq. (1) in the main paper

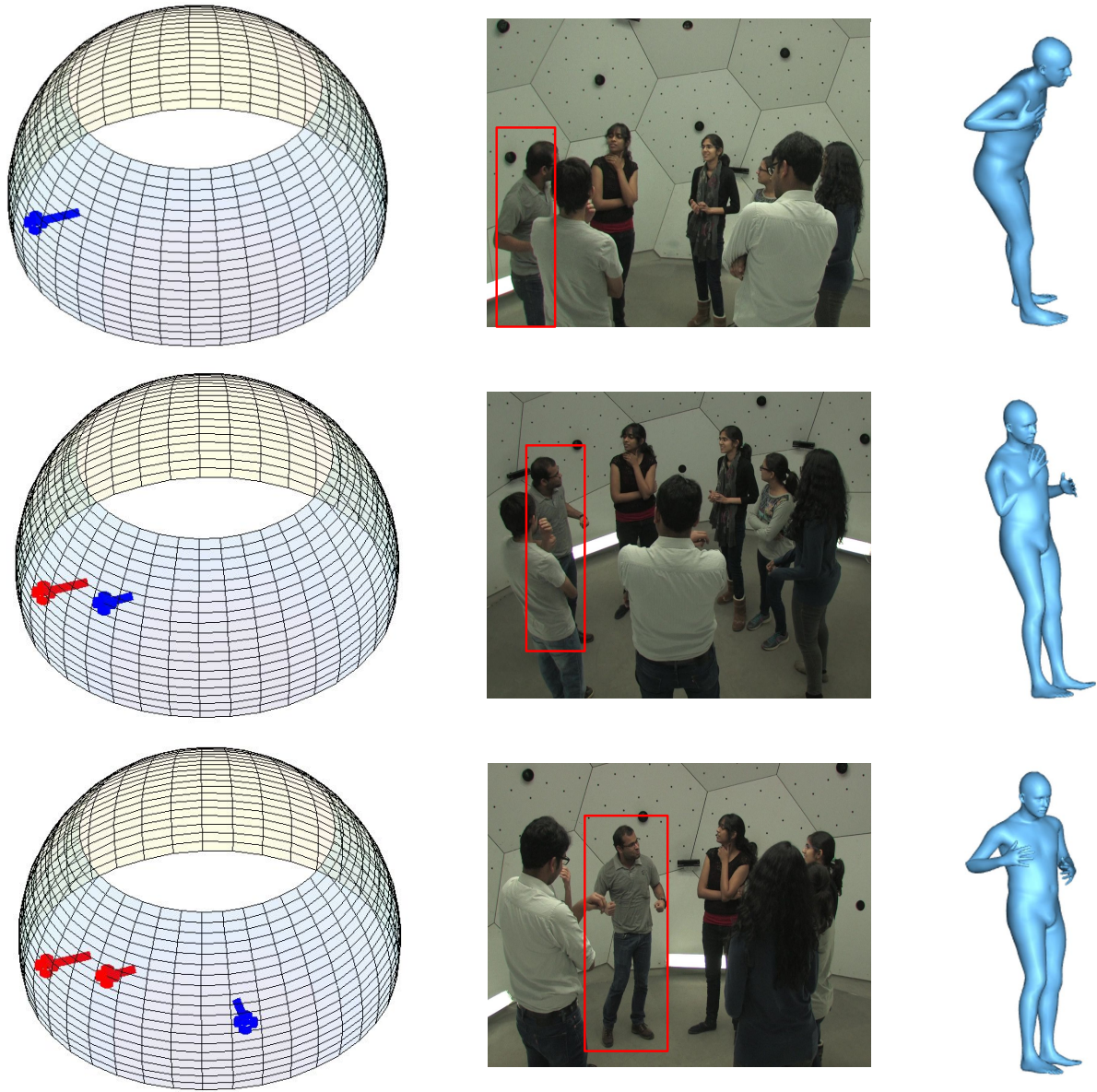

Figure 2: Visualization of how Pose-DRL performs single-target reconstruction on an active-view (set of viewpoints for a time-freeze) for a *Mafia* test scene. In this case the agent sees three viewpoints prior to automatically continuing to the next active-view. The reconstruction error reduces from 168 to 107 mm/joint. Left: Viewpoints seen by the agent, where blue marks the current viewpoint (camera) and red marks previous viewpoints. Note that the initial camera was given randomly. Middle: Input images associated to the viewpoints, also showing the detection bounding box of the target person in red – detections for the other people are left out to avoid visual clutter. Right: SMPL visualizations of the partially fused poses, cf. (1). The target person is only partially visible in the initial viewpoint, and the associated pose estimate is inaccurate with the reconstruction incorrectly tilting forward. As the agent visits more viewpoints, the stance of the reconstruction becomes straighter and more correct. The person is fully visible in the final viewpoint, and the associated final fused estimate is accurate and plausible.

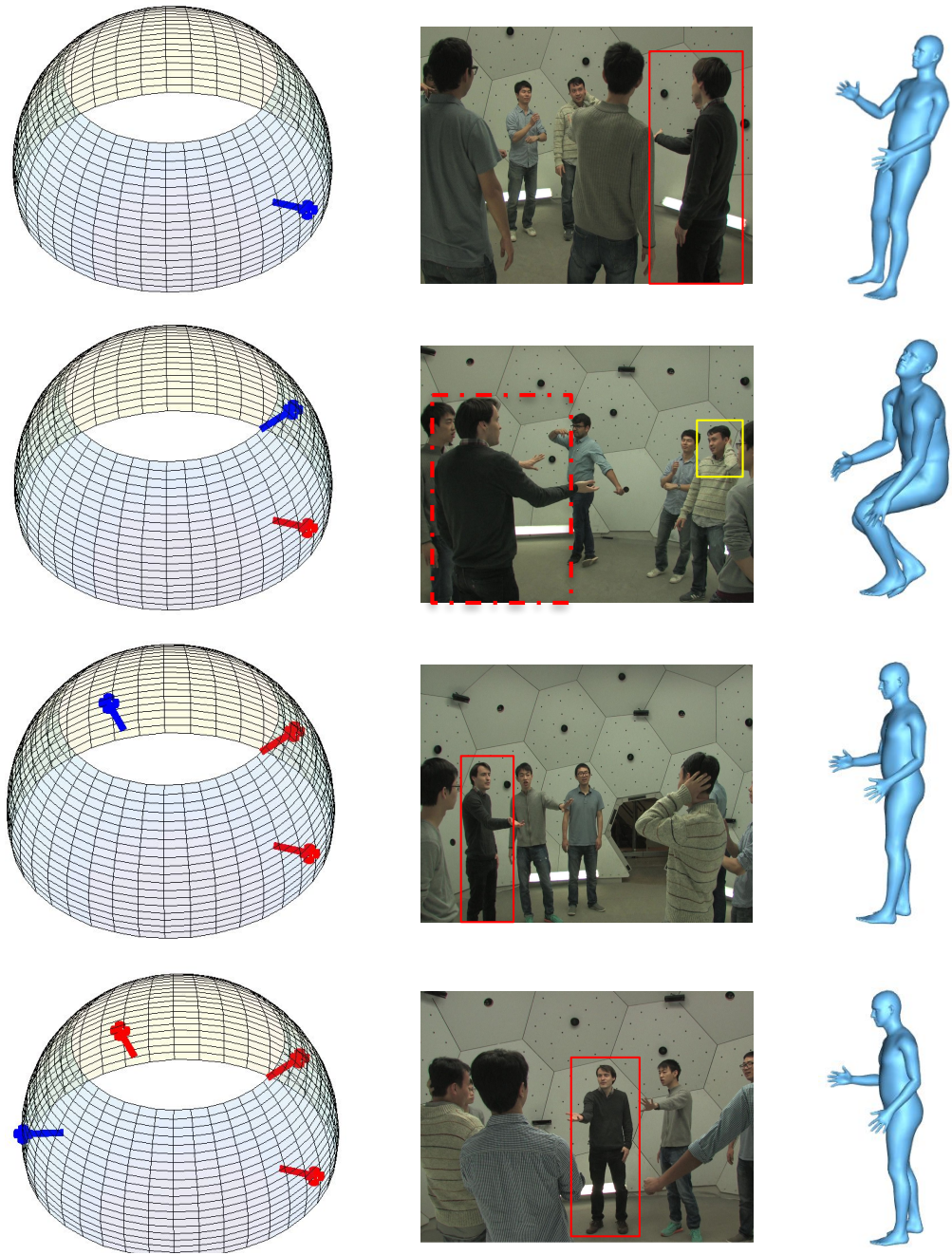

Figure 3: Visualization of how Pose-DRL performs single-target reconstruction on an active-view (set of viewpoints for a time-freeze) for an *Ultimatum* test scene, where in this case the detection and matching is incorrect for the second viewpoint. Left: Viewpoints seen by the agent, where blue marks the current viewpoint (camera) and red marks previous viewpoints. Note that the initial camera was given randomly. Middle: Input images associated to the viewpoints, also showing the detection bounding box of the target person in red – detections for the other people are left out to avoid visual clutter. In the second viewpoint with the incorrect detection and matching, the target person is indicated with a dashed red bounding box, and the incorrect detection used in the pose fusion is shown in yellow. Right: SMPL visualizations of the partially fused poses, cf. (1). The target person is viewed from a suboptimal direction in the first viewpoint, causing the associated pose estimate to be incorrectly tilted. As the agent moves to the next viewpoint to get a better view of the person, the underlying detection and matching system suggests an incorrect detection to feed the pose estimator, which causes the fused estimate to deteriorate severely. However, the agent is able to remedy this by selecting two more good and diverse viewpoints where the target is clearly visible, yielding a considerably better fused pose estimate. In this example the agent sees four viewpoints prior to automatically continuing to the next active-view. The reconstruction error reduces from 149 to 119 mm/joint.

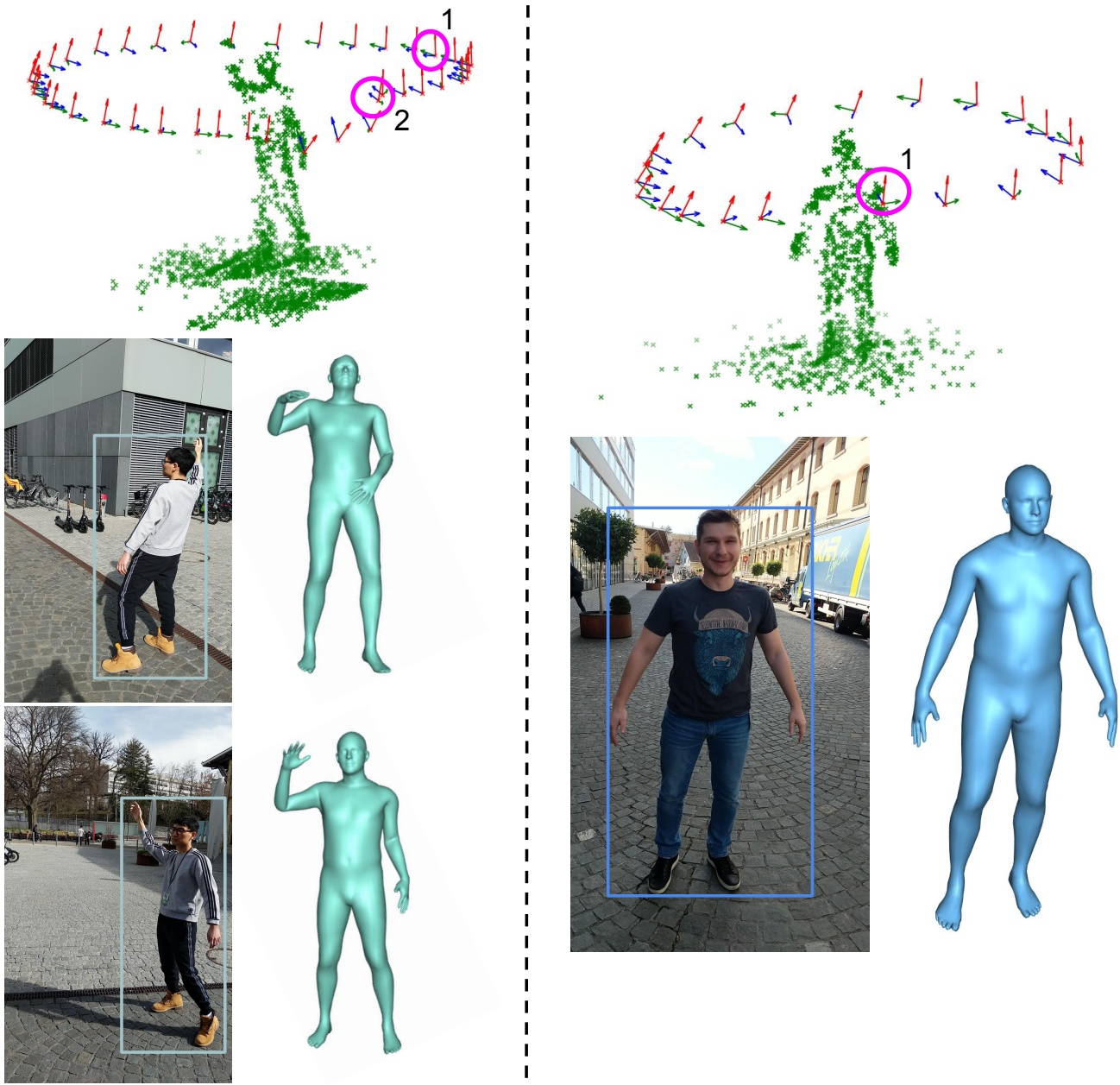

Figure 4: People standing in various poses, captured with a smartphone camera from different viewpoints. Note that this data is significantly different from that obtained from Panoptic, with more challenging outdoor lighting conditions, human-imposed errors from holding and directing the smartphone camera, etcetera. We show two visualization of how Pose-DRL operates in different scenarios. Pose-DRL was *not* re-trained on this data; we use the same model weights as for producing the results in the main paper. In each scenario we also show the 3d configuration of the scene, as well as which viewpoints are selected by the agent and in which order (pink circles). Left: In this example the agent sees two views before terminating viewpoint selection. The initial randomly given viewpoint produces a pose estimate where the arms are not accurate, which is corrected for in the second and final viewpoint. Right: The agent receives a very good initial viewpoint and decides to terminate viewpoint selection immediately, producing an accurate pose estimate. See § 4.1 for more details about these visualizations.



\end{document}